\documentclass[letterpaper, 10 pt, conference]{ieeeconf}  

\IEEEoverridecommandlockouts                              

\usepackage[hypertexnames=false,hidelinks]{hyperref}
\usepackage[all]{hypcap}
\usepackage[noadjust]{cite}
\usepackage{graphics} 
\usepackage{graphicx,amsfonts,caption}
\usepackage{amsmath,dsfont} 
\usepackage{array}
\usepackage{booktabs}
\usepackage{subfig} 
\usepackage{float}
\usepackage{xcolor}
\usepackage{tikz}
\usepackage[capitalize]{cleveref}

\title{\LARGE \bf
STEAD: Spatio-Temporal Efficient Anomaly Detection\\for Time and Compute Sensitive Applications 

}

\author{Andrew Gao and Jun Liu
\thanks{The authors are with Department of Computer Engineering,
        San Jose State University, 1 Washington Sq, San Jose, CA 95192 USA
     (e-mail: andrew.gao@sjsu.edu, junliu@sjsu.edu).}%
}

\begin{document}

\maketitle
\thispagestyle{empty}

\begin{abstract}

This paper presents a new method for anomaly detection in automated systems with time and compute sensitive requirements, such as autonomous driving, with unparalleled efficiency. 
As systems like autonomous driving become increasingly popular, ensuring their safety has become more important than ever. Therefore, this paper focuses on how to  \emph{quickly} and \emph{effectively} detect \emph{various} anomalies in the aforementioned systems, with the goal of making them safer and more effective. Many detection systems have been developed with great success under spatial contexts; however, there is still significant room for improvement when it comes to temporal context. While there is substantial work regarding this task, there is minimal work done regarding the efficiency of models and their ability to be applied to scenarios that require \emph{real-time} inference, i.e., autonomous driving where anomalies need to be detected the moment they are within view. To address this gap, we propose STEAD (Spatio-Temporal Efficient Anomaly Detection), whose backbone is developed using (2+1)D Convolutions and Performer Linear Attention, which ensures computational efficiency without sacrificing performance. When tested on the UCF-Crime benchmark, our base model achieves an AUC of 91.34\%, outperforming the previous state-of-the-art, and our fast version achieves an AUC of 88.87\%, while having 99.70\% less parameters and outperforming the previous state-of-the-art as well. The code and pretrained models are made publicly available at \url{https://github.com/agao8/STEAD}.

\end{abstract}

\section{Introduction}

The rapid proliferation of automated systems powered by artificial intelligence and machine learning has underscored the critical need for robust anomaly detection mechanisms. In those applications,
the ability to detect anomalous events in real time is paramount to ensuring safety, security, and operational continuity. As such, anomaly detection (AD) and video anomaly detection (VAD) systems, which identify deviations, abnormalities, and anomalies within normal patterns has emerged as a pivotal research area in autonomous driving. As image classification models become increasingly powerful, it is no surprise that anomaly detection models, which perform a sub-task of image classification, have seen great success on spatial scenarios without temporal contexts, with numerous models reaching near perfect results on popular datasets like the MTtec Anomaly Detection Dataset \cite{bergmann2019mvtec}. 

\emph{``What is an anomaly in autonomous driving?''} 

\begin{figure}[!t]
  \centering
  \includegraphics[width=1\linewidth]{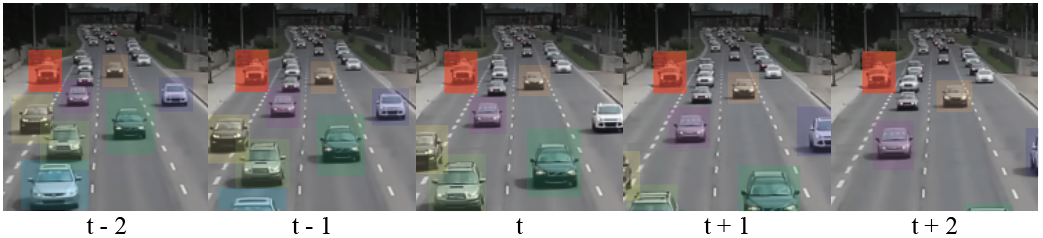}
  \caption{\emph{Normal vs. abnormal}: without any temporal context information, the frame at time-step $t$ seems completely \emph{normal}. However, when considering the frames at other time-steps, it can be seen that there is a vehicle in the top left, highlighted in red, that is not moving despite being on the road, while all the other vehicles are moving as expected. As such, the entire video should be labeled to contain an \emph{abnormal} event, yet that conclusion cannot be reached from any 1 frame of the video.}
  \label{fig:highway}
\end{figure}

In the context of autonomous driving, the definition of what an ``anomaly'' is is relative to the definition of ``normal'', and unfortunately, what is considered ``normal'' is also dependent on the scenario. Yet, while many previous works operate solely under spatial contexts, it is unreasonable to build an understanding of what is ``normal'' in a completely unseen environment without a temporal context in the context of autonomous driving. For example, in Figure \ref{fig:highway} which depicts some frames from a video of an active road \cite{ythighway}, just by evaluating the video at frame $t$, it is impossible to accurately conclude that there is an anomaly present in the frame. However, once the surrounding frames are brought into consideration through a temporal context, it can be seen that all the vehicles are moving from the top of the frame to the bottom, except for the vehicle highlighted in red in the top left. It is apparent that the location of the vehicle highlighted in red should not be immobile, and therefore it is to be considered an anomaly, yet that conclusion could not be reached just by evaluating the video at time frame $t$. Despite scenarios like this, the challenges and benefits presented by the temporal dimension inherent to video data remain inadequately addressed to ensure a safety driving environment.

Traditional approaches to identify abnormal behavior in autonomous driving relied on handcrafted features and statistical models, to model normal behavior and flag deviations. However, these methods often falter in complex, dynamic environments due to their limited capacity to capture high-dimensional spatiotemporal patterns. The advent of deep learning revolutionized the field, with Convolutional Neural Networks being the dominant force behind most tasks within computer vision and more recently, Vision Transformers, which are gaining popularity and success. Despite these advancements, existing models frequently prioritize accuracy at the expense of computational efficiency, rendering them impractical for real-time applications, where their high computational cost and memory footprint hinder deployment in latency-sensitive scenarios.

A critical gap in the field lies in balancing \emph{efficiency} with \emph{performance}. Many state-of-the-art methods, such as transformer-based architectures, suffer from quadratic complexity in attention mechanisms or excessive parameter counts, limiting scalability. Furthermore, the reliance on inefficient feature extractors like I3D \cite{carreira2018quovadisactionrecognition} exacerbates these challenges. Recent efforts to address efficiency, such as Linformer \cite{wang2020linformerselfattentionlinearcomplexity} and Performers \cite{choromanski2022rethinkingattentionperformers}, demonstrate the potential of linear-time attention mechanisms but have not been fully exploited in identifying abnormal behavior in autonomous driving scenarios.

This paper introduces STEAD (Spatio-Temporal Efficient Anomaly Detection), a novel architecture designed to bridge the efficiency-performance divide in monitoring and identifying abnormal behavior in autonomous driving. Moreover, we developed two different versions: STEAD-Base, which focuses on achieving high performance, and STEAD-Fast, which prioritizes efficiency without sacrificing performance too much. In summary, our contributions are threefold:
\begin{itemize}
    \item \emph{State-of-the-art Performance}: STEAD-Base and STEAD-Fast achieve respective AUC scores of 91.34\% and 88.87\%, becoming the new state-of-the-art on the benchmark \cite{sultani2019realworldanomalydetectionsurveillance}.

    \item \emph{Real-Time Processing}: STEAD-Base and STEAD-Fast leverage decoupled 3D convolutions and linear-time attention mechanisms to reduce computational costs while preserving discriminative power, surpassing previous the state-of-the-art in performance despite having a significantly respective 71.85\% and 99.70\% reduction in total parameter counts.



    \item \emph{High-Performance Feature Extraction}: By leveraging the X3D model \cite{feichtenhofer2020x3dexpandingarchitecturesefficient} as a lightweight yet powerful alternative to I3D, we achieve superior accuracy with significantly fewer parameters.
\end{itemize}


These advancements position our framework as a practical solution for \emph{real-time} applications, where computational efficiency and accuracy are equally critical. By addressing the dual challenges of temporal context modeling and resource constraints, this work advances the frontier of abnormal behavior detection in autonomous driving toward deployable, scalable systems.

\section{Related Work}
\subsection{Video Anomaly Detection}

Detecting abnormal behavior in autonomous driving has garnered significant attention in recent years due to its importance in ensuring safety. This method can also be applied in applications in surveillance, security, and behavior analysis. Early approaches often relied on handcrafted features and statistical models. For instance, Mahadevan et al. introduced a robust framework that utilizes Gaussian Mixture Models (GMM) for background modeling, enabling the detection of deviations from learned normal patterns in videos \cite{5539872}.

In recent years, deep learning has revolutionized the field, with researchers such as Ahmed et al., who proposed a deep learning-based framework that employs Convolutional Neural Networks to automatically extract features from video data, achieving state-of-the-art results in detecting anomalies at the time \cite{10072765}. Another notable approach that has shown promise is the use of generative models. Zenati et al. employed Generative Adversarial Networks (GANs) for anomaly detection, demonstrating that the model could effectively learn the distribution of normal events and highlight anomalies based on reconstruction errors \cite{8594897}. Additionally, Liu et al. proposed a two-stream network architecture that separately processes spatial and temporal information, further enhancing detection accuracy \cite{liu2023real}. More modern approaches seek to exploit the success of Vision Transformers for anomaly detection tasks, such as the TransAnomaly model proposed by Yuan et al. \cite{9525368}. Furthermore, researchers also sought to build Siamese Networks on top of vision transformers to learn feature representations. For instance, Chen et al. built a Siamese Network that utilizes a magnitude contrastive loss to encourage feature separability \cite{chen2022mgfnmagnitudecontrastiveglanceandfocusnetwork}. 

\subsection{Feature Extraction}


Extracting spatio-temporal features is critical in identifying abnormal behavior in autonomous driving. Transfer learning via feature extraction has become a cornerstone of modern machine learning due to its efficiency and effectiveness, with previous works on this topic primarily using the Inflated 3D ConvNet (I3D) model as their feature extractor \cite{zhou2023batchnormbasedweaklysupervisedvideo, chen2022mgfnmagnitudecontrastiveglanceandfocusnetwork, pu2024learningpromptenhancedcontextfeatures, ali2020selfsupervisedrepresentationlearningvisual}. The I3D model sought to addresses the challenge of spatiotemporal feature learning in video recognition by extending successful 2D convolutional architectures into 3D \cite{carreira2018quovadisactionrecognition}. To accomplish this, the model ``inflated'' 2D convolutional kernels into 3D by replicating and pooling them along the temporal dimension \cite{carreira2018quovadisactionrecognition}. The model also incorporates a two-stream design, both RGB and optical flow, to enhance feature modeling \cite{carreira2018quovadisactionrecognition}.

However, the I3D model suffers from a high computational cost in both memory requirements as well as processing power, making it impracticable for real-time inference applications. Alternatively, while taking note of the trade-off between computational efficiency and accuracy, the X3D model sought to progressively expand a minimal 2D backbone along the temporal duration, frame rate, spatial resolution, channel width, network depth, and bottleneck width axes \cite{feichtenhofer2020x3dexpandingarchitecturesefficient}. Throughout training, at each step, the methodology would instruct the model to expand each axis, in isolation to each other, and then select a single axis to keep expanded for future steps based on which axis expansion achieved the greatest computation to accuracy trade-off after training and validation. This continues until the model reaches a desired computational cost \cite{feichtenhofer2020x3dexpandingarchitecturesefficient}. As a result, the X3D family of models boasts a notably higher top-1 accuracy on popular datasets such as Kinetics-400 \cite{kay2017kineticshumanactionvideo}, while requiring less than a quarter of the parameters used by the I3D model. Consequently, our work will utilize the X3D model to extract video features over the I3D model used by previous works.


\subsection{Vision Transformers}
Convolutional Neural Networks (CNNs) have been extensively studied and applied in various domains, particularly in computer vision tasks. 
Since then, countless architectures and been proposed and tested in the pursuit of more powerful and efficient models. One particular method useful for video processing is the (2+1)D Convolution, which applies a spatial 2D convolution proceeded by a temporal 1D convolution over directly applying a 3D convolution, which reduces computational complexity and overfitting \cite{tran2018closerlookspatiotemporalconvolutions}. However, while CNNs have been at the forefront for many years, vision transformer models are becoming evermore prevalent, with their performance matching or even exceeding that of CNNs.
 
Transformers based on self-attention mechanisms were initially designed for and utilized in natural language processing tasks such as machine translation \cite{vaswani2023attentionneed}. Since then, they have become the dominant method in the vast majority of natural language processing tasks, consistently outperforming other methods for state-of-the-art performance. 

Inspired by their success in natural language processing, researchers sought to expand the application of transformers to 2D computer vision tasks, kick-started by the introduction of the Vision Transformer, which applied the transformer architecture directly on a sequence of image patches for image classification \cite{dosovitskiy2021imageworth16x16words}. This effort proved to successful as transformer based models went on to utilized in a variety of computer vision tasks such as image classification, action recognition, object detection, image segmentation, image and scene generation \cite{liu2021swintransformerhierarchicalvision, 10655590, carion2020endtoendobjectdetectiontransformers, strudel2021segmentertransformersemanticsegmentation, ramesh2021zeroshottexttoimagegeneration}. 

However, the original transformer architecture is limited by a quadratic increase in both time and space complexity with image size, making it a bottleneck for large-scale applications. In response to this issue, researchers proposed models such as Linformer \cite{wang2020linformerselfattentionlinearcomplexity}, which lowered computational costs to $\mathcal O (nk)$, where $k$ is a lower-rank projection of $n$, the length of the sequence, and the Performer \cite{choromanski2022rethinkingattentionperformers}, which used Fast Attention Via Orthogonal Random Features to approximate the kernel in linear time. Our work explores these techniques as we utilize their efficiency for higher dimensional video data.

\subsection{Differentiating Abnormal from Normal Behavior}

\begin{figure}[!t]
  \centering
  \includegraphics[width=1\linewidth]{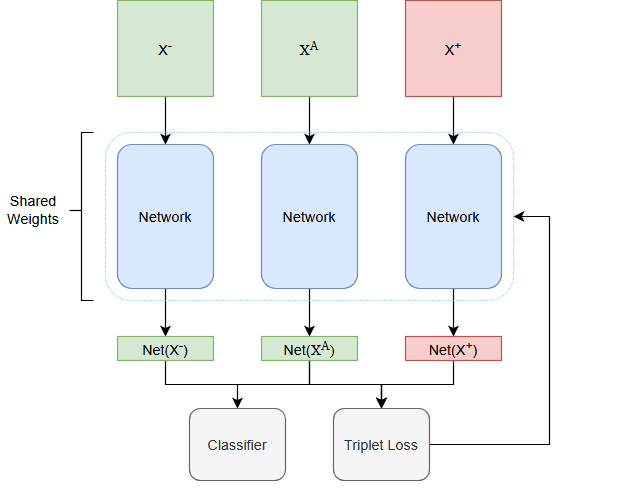}
  \caption{An overview of the triplet network. Three inputs are passed through the network and independently processed to obtain feature embeddings. The inputs consist of an anchor (typically the same class as the negative) ($\mathbf x^A$), a negative($\mathbf x^-$), and a positive sample ($\mathbf x^+$). Once feature embeddings have been obtained, the results are passed to a classifier to construct a classification score and to a Triplet Loss function to compute loss.}
  \label{fig:triplet}
\end{figure}

Siamese networks were first introduced for signature verification, which introduced the concept of training dual neural networks to assess the similarity between pairs of inputs \cite{10.5555/2987189.2987282}. However, they were not popularized until much later with the advent of deep learning, with the publications made by Koch and Schroff et al., who developed models for one-shot image recognition and face recognition respectively \cite{koch2015siamese, Schroff_2015}. Koch's model emphasized the significance of similarity learning through the usage of a contrastive loss function \cite{koch2015siamese}. On the other hand, Schroff et al. improved upon the contrastive loss function by developing triplet loss, which incorporated an additional anchor input in the calculation of similarity, which allowed the model to explicitly learn the relative distances between embeddings, ensuring that the similarity between the anchor and positive was minimized relative to that between the anchor and negative \cite{Schroff_2015}. This concept would then be refined by Hoffer Ailon \cite{hoffer2015deep} who proposed the Triplet Network as an extension of the Siamese Network that utilizes a trio of neural networks instead of the Siamese Network's traditional use of two as shown in figure \ref{fig:triplet}.

This family of networks has since been proven to be highly successful in tasks where similarity metrics are a core component or can be exploited if they are not, such as image similarity ranking, sentence similarity learning, and sentence embedding \cite{wang2014learning, 9423550, reimers2019sentence}.

\begin{figure*}[!t]
  \centering
  \includegraphics[width=0.9\linewidth]{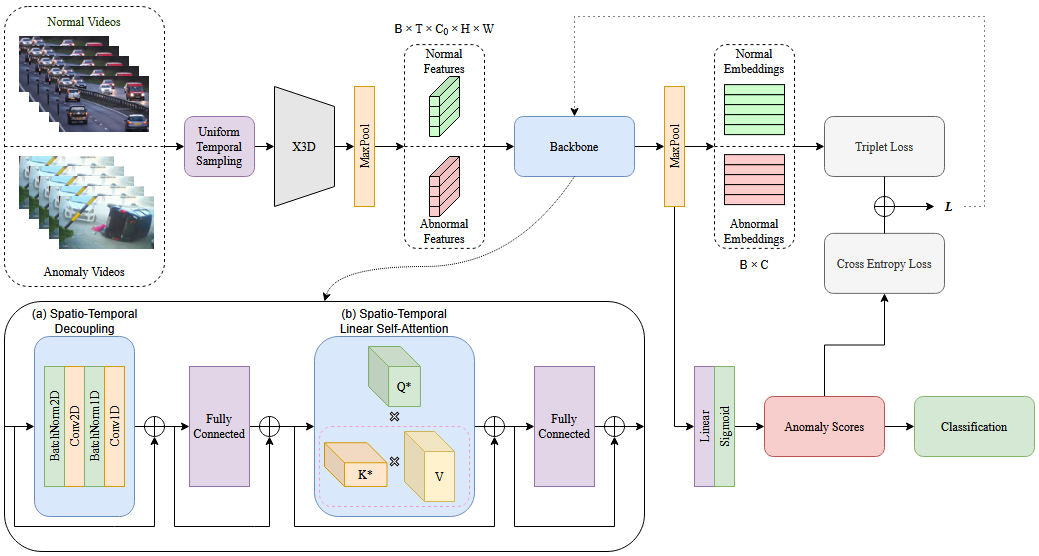}
  \caption{An overview of our proposed network architecture. Input videos are first split into clips via Uniform Temporal Sampling and then each clip has its features extracted using X3D; these features of each clip for a given video are then pooled together to obtain a single feature representation for the whole video. Next, (a) decouples the spatial and temporal dimensions to enhance the individual feature dimensions. Then (b) captures long range relations between the temporal and spatial dimensions. Final video embeddings are obtained after a pooling across both the temporal and spatial dimensions. Finally, a loss is computed by combining the triplet loss of the embeddings and the cross entropy loss of the classifications.}
  \label{fig:arch}
\end{figure*}

\nocite{ythighway2}

\section{Abnormal Behavior Detection}

In this section, we will first introduce our architecture for identifying abnormal behavior in autonomous driving, and then delve into the different parts, such as feature extraction, in the following sections.

\subsection{Overall Architecture}

The overall architecture of our model for abnormal behavior detection in autonomous driving is shown in Figure \ref{fig:arch}. Before being used by our model, videos are first transformed with a center crop and then uniformly sampled, splitting an individual video into clips, which then have their features extracted by the X3D model. The resulting per clip features of shape $C \times T \times H \times W$, where $C, T, H$, and $W$ denote the number of channels, temporal dimension, height, and width of the features respectively, are then pooled together to obtain the most prominent features for each video. From here, the video features are passed as input to our model where they are processed by a decoupled spatio-temporal feature enhancer (a) to refine the features in preparation for the spatio-temporal attention block (b) where long range connections between features are modeled using linear self-attention. After the features of each video are processed by each block, \textit{T}, \textit{H}, \textit{W}, and {d} are pooled to result in a final video feature embedding of size \textit{d}, where $d$ is the size of the channel dimension of the last layer. Instead of evaluating our model only based on classification metrics, e.g., cross entropy, we also employ a triplet loss as shown in equation~\eqref{eq:loss_triplet}, designed to maximize numerical separability between features of normal and abnormal samples and will be illustrated later in this section.


\begin{figure}[!t]
  \centering
  \includegraphics[width=0.6\linewidth]{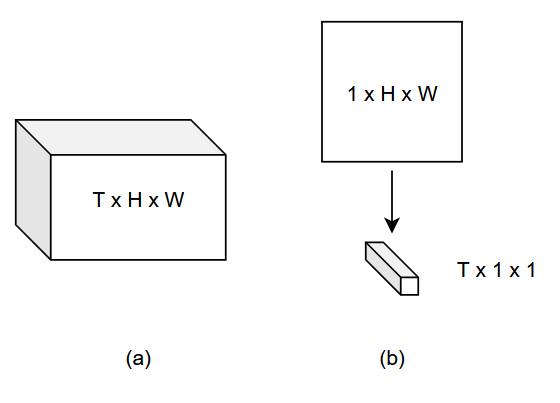}
  \caption{An illustration of spatiotemporal decoupling for an input of shape $T \times H \times W$, where $T$ denotes the temporal dimension and $H$ and $W$ denote the spatial dimensions. Channels are not shown for simplicity. (a) A traditional 3D convolution with a 3D filter. (b) Spatiotemporal decoupled convolution where convolution is split into a 2D spatial convolution followed by a 1D temporal convolution.}
  \label{fig:2plus1}
\end{figure}

\subsection{Abnormal Behavior Detection}


Our decoupled spatio-temporal feature enhancer follows the (2 + 1)D Convolution architecture \cite{tran2018closerlookspatiotemporalconvolutions}, which reduces computational complexity over directly performing 3D convolutions. As illustrated in Figure \ref{fig:2plus1}, our feature enhancer consists of a (2 +1)D Convolution, which is followed by a Feed Forward Network containing two fully connected layers with GELU (Gaussian Error Linear Unit) non-linearity in between. A LayerNorm is applied before each module and a residual connection is added after each module.

The spatio-temporal attention block utilizes the Performer attention architecture, promoting linear efficiency over the original transformer \cite{choromanski2022rethinkingattentionperformers}. The original attention proposed by \cite{vaswani2023attentionneed} is shown below in equation \eqref{eq:attention}, where $Q, K, V \in \mathbb R^{L \times C}$ are the query, key, and value, $L = (T \times H \times W)$. 

\begin{equation}
    \text{Attention(K, Q, V)} = \frac{AV}{D},
    \label{eq:attention}
\end{equation}
where
\begin{equation*}
    A = \exp\left(\frac{QK^T}{\sqrt{C}}\right),
\end{equation*}
and 
\begin{equation*}
    D = \text{diag}(A \cdot \mathds 1_L).
\end{equation*}

The variable $A \in \mathbb R^{L \times L}$ computes an attention matrix from $Q$ and $K$ such that $AV$ normalized by $D$, defines the attention operation, where $D \in \mathbb R^{L \times L}$ is a normalization matrix that results in a softmax operation once applied. The variable $\mathds 1_L$ denotes a vector of length $L$ containing all 1s. However, this suffers from a quadratic computation cost with respect to $L$ when calculating $\text{exp}(\frac{QK^\top}{\sqrt{C}})$. Performer attention solves this by formulating a softmax kernel approximation as follows:

\begin{equation}
    \exp\left(\frac{QK^T}{\sqrt{C}}\right) \approx \phi(Q) \phi(K)^\top,
    \label{eq:approx}
\end{equation}
and rearranging the attention calculation as:
\begin{equation}
    \begin{split}
        \widehat{\text{Attention(K, Q, V)}} \approx \frac{\phi(Q) (\phi(K)^\top V)}{\hat{D}},
    \end{split}
    \label{eq:attention_approx}
\end{equation}
where 
\begin{equation*}
    \hat{D} = \phi(Q) (\phi(K)^\top \mathds 1_L),
\end{equation*}
and $\phi: \mathbb R^{C} \xrightarrow{} \mathbb R^{C'}$ is a feature map and $d' > d$ . By mapping $Q$ and $K$ to a higher dimension $d'$, accurate approximations of the softmax kernel are made while rearranging the calculation reduces the computation cost to be linear with respect to $(T \times H \times W)$.

The Performer attention is followed by a Feed Forward Network containing two fully connected layers with GELU non-linearity in between. A LayerNorm is applied before each module and a residual connection is added after each module.



Since detecting abnormal behavior in autonomous driving using videos is a binary-class classification problem, it is natural to use binary cross entropy loss to compare the predicted probability scores of each video against the ground truths. However, to encourage feature separability between normal and abnormal samples, we also utilize a triplet loss function as shown below, where the normal video embeddings serve as both the negative and the anchor.
\begin{equation}
    \begin{split}
& \mathcal{L}_t (\mathbf e _i) \\
= & \frac{1}{N}\sum_{i=0}^{N} \max_{\substack{j \neq i \\ j=0,\ldots,N}} d(\mathbf e^n_i, \mathbf e^n_j) + \\ 
& \frac{1}{N}\sum_{i=0}^{N} \max\left(0, M - \min_{\substack{j \neq i \\ j=0,\ldots,N}} d(\mathbf e^n_i, \mathbf e^a_j)\right),
    \end{split}
\label{eq:loss_triplet}
\end{equation}
where $\mathbf e_i = \{\mathbf e^n_i, \mathbf e^a_j | i,j = 1,\ldots, N\}$ and $\mathbf e^n_i, \mathbf e^a_j \in \mathbb R^{d}$ refer to the final feature embeddings of the \emph{normal} and \emph{abnormal} from $i$th and $j$th sample respectively. Every batch is of size $2N$, with $N$ normal videos and $N$ abnormal videos. $d: \mathbb R^d \times \mathbb R^d \mapsto \mathbb R$ is the Euclidean distance. A maximum margin $M$ of the distance between normal and abnormal feature embeddings is enforced to restrict the model from optimizing purely for maximizing the distance between normal and abnormal embeddings. Embedding distance function $D$ is defined as Euclidean distance. 

Combining this triplet loss function with binary cross-entropy loss
\begin{equation*}
\mathcal{L}_c (\mathbf e_i) = - \frac{1}{2N} \sum_{i=1}^{2N} \left[ y_i \log(\hat{y}_i) + (1 - y_i) \log(1 - \hat{y}_i) \right],
\end{equation*}
where $y_i$ is the ground truth of $\mathbf e_i$ and $\hat y_i$ is the prediction of $\mathbf e_i$, we can get an overall loss function $\mathcal{L}$
\begin{equation*}
\mathcal{L} (\mathbf e_i) = \mathcal L_c (\mathbf e_i) + \lambda \cdot \mathcal L_t (\mathbf e_i),
\label{eq:loss}
\end{equation*}
where $i=1, \ldots, 2N$ and $\lambda \in \mathbb R$ is controls the weight of the triplet loss term.

\begin{table}[!t]{\small
\begin{center}
\caption{Comparison of AUC-ROC and feature extractors with existing works. Our models, STEAD and its variants, vastly outperform \emph{all} previous works. 
}
\begin{tabular}{ccc}
\toprule
Method & Feature Extractor & AUC-ROC(\%) \\
\midrule
Sultani et al. \cite{sultani2019realworldanomalydetectionsurveillance} & C3D & 75.41\\
Sultani et al. \cite{sultani2019realworldanomalydetectionsurveillance} & I3D & 77.92\\
MIST \cite{feng2021mistmultipleinstanceselftraining} & I3D & 82.30\\
CLAWS \cite{zaheer2021clawsclusteringassistedweakly} & C3D & 83.03\\
RTFM \cite{tian2021weaklysupervisedvideoanomalydetection} & VideoSwin & 83.31\\
RTFM \cite{tian2021weaklysupervisedvideoanomalydetection} & I3D & 84.03\\
WSAL \cite{Lv_2021} & I3D & 85.38\\
S3R \cite{WuHCFL22} & I3D & 85.99\\
MGFN \cite{chen2022mgfnmagnitudecontrastiveglanceandfocusnetwork} & VideoSwin & 86.67\\
PEL \cite{pu2024learningpromptenhancedcontextfeatures} & I3D & 86.76\\
UR-DMU \cite{zhou2023dualmemoryunitsuncertainty} & I3D & 86.97\\
MGFN \cite{chen2022mgfnmagnitudecontrastiveglanceandfocusnetwork} & I3D & 86.98\\
BN-WVAD \cite{zhou2023batchnormbasedweaklysupervisedvideo} & I3D & 	87.24\\
\textbf{STEAD-Fast (Ours)} & \textbf{X3D} & \textbf{88.87}\\
\textbf{STEAD-Base (Ours)} & \textbf{X3D} & \textbf{91.34}\\
\bottomrule
\end{tabular}
\vspace*{-4mm}
\label{table:ucf-results}
\end{center}}
\end{table}

\begin{table}[!t]{\small
\begin{center}
\caption{Comparison of AUC-ROC and number of parameters with existing works on the benchmark. STEAD-Fast outperforms \emph{all} previous works while having significantly less parameters. STEAD-Base achieves even greater performance improvements.
}
\begin{tabular}{ccc}
\toprule
Method & AUC-ROC(\%) &  \# of Parameters\\
\midrule
RTFM \cite{tian2021weaklysupervisedvideoanomalydetection} & 84.03 & 24.72M\\
WSAL \cite{Lv_2021} & 85.38 & 1.98M\\
S3R \cite{WuHCFL22} & 85.99 & 325.74\\
PEL \cite{pu2024learningpromptenhancedcontextfeatures} & 86.76 & 1.21M\\
MGFN \cite{chen2022mgfnmagnitudecontrastiveglanceandfocusnetwork} & 86.98 & 26.65M\\
BN-WVAD \cite{zhou2023batchnormbasedweaklysupervisedvideo} & 87.24 & 5.79M\\
\textbf{STEAD-Fast (Ours)} & \textbf{88.87} & \textbf{17,441}\\
\textbf{STEAD-Base (Ours)} & \textbf{91.34} & \textbf{1.63M} \\
\bottomrule
\end{tabular}
\vspace*{-4mm}
\label{table:size_comparison}
\end{center}}
\end{table}

\section{Experiments}
For analysis, we conduct multiple experiments on the benchmark \cite{sultani2019realworldanomalydetectionsurveillance}, which consists of weakly-labeled normal and abnormal videos in different scenarios. Especially, we test the peroformance of our models on different scenarios related to autonomous driving. 

\subsection{Experiments Settings}

The dataset contains $128$ hours of surveilance video footage split between 1900 videos, 940 of which are labeled as normal and 960 as abnormal. 140 normal videos and 150 abnormal videos are reserved for evaluation, for a totaling testing set size of 290. The abnormal videos consist of different types of anomalies, such as road accidents, explosions, assaults, etc.


The performance was evaluated under 2 architectures, STEAD-Base and STEAD-Fast. For both architectures, the extracted video features are of shape $(C, T, H, W) = (192, 16, 10, 10)$, where \textit{C}, \textit{T}, \textit{H}, and \textit{W} denote the number of channels, number of frames, height, and width of the features respectively. Each architecture was formed with a single featuer enhancer block followed by an attention block. For the base architecture, the feature enhancer and attention blocks had respective channel dimensions of 192 and 128, both with a depth of 3. For the Fast architecture, the blocks both had channel dimensions of 32 and a depth of 1. Both models were trained with AdamW \cite{loshchilov2019decoupledweightdecayregularization}, a weight decay of 0.2, a cosine decay learning rate scheduler, and $M$ = 100. 

\begin{figure}[!t]
    \centering
    \subfloat[4 clips from a video containing normal events (1 \& 2) and abnormal events (3 \& 4). Clips 1 and 2 showcase the scenario \emph{before} any abnormal event has occurred. Clip 3 and 4 show an abnormal road accident event in which a motorbike collides with a vehicle.]{\includegraphics[width=0.9\linewidth]{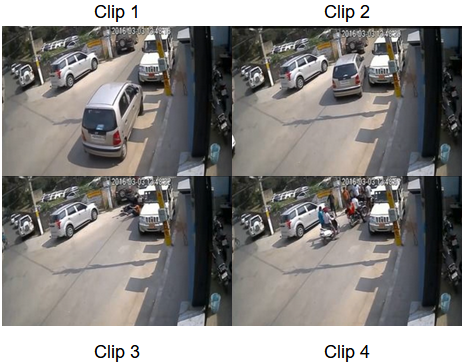}}%
    \label{fig:ex_clip}
    \vfil
    \subfloat[UMAP embedding of each clip.]{\includegraphics[width=0.8\linewidth]{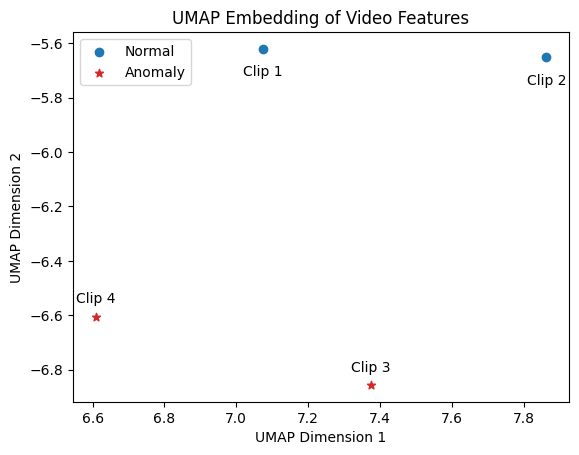}%
    \label{fig:ex_embed}}
    \caption{Example of an anomalous video being split into individual 4 clips and the resulting UMAP embedding of each clip. Each clip from (a) is passed through the model independently (without pooling from feature extraction) and the resulting UMAP embeddings are shown in (b). }
    \label{fig:example}
\end{figure}

We then compare the results of our method with previous state-of-the-arts. Following previous works, we adopt Area Under the Curve (AUC) under the Receiver Operating Characteristic (ROC) as the evaluation metric. An example result of our model is shown in figure \ref{fig:example}, where a video of a road accident anomaly has been split into 4 separate clips. Clips 1 and 2 do not contain any anomaly, while clips 3 and 4 contain a road accident anomaly in which a motorbike and a vehicle collide, located in the upper right region of the video. The model makes correct classifications for all 4 clips and is shown to be able to linearly separate the clip embeddings.

\subsection{Results Analysis}


\begin{figure}[!t]
  \centering
  \includegraphics[width=0.9\linewidth]{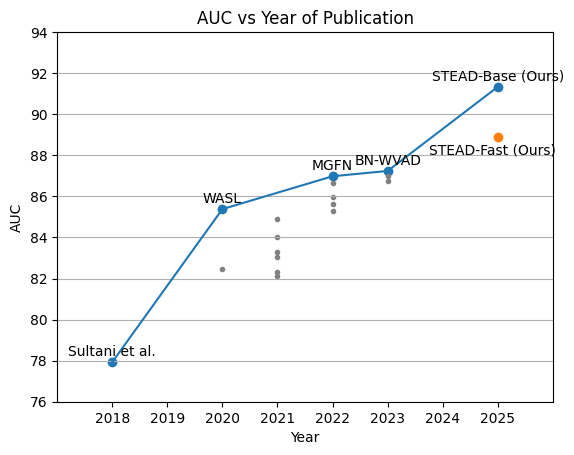}
  \caption{State-of-the-art AUC performance over time on the benchmark. Both STEAD-Base and STEAD-Fast obtain significantly higher results and become the new state-of-the-art.}
  \label{fig:auc_time}
\end{figure}

\begin{figure}[!t]
    \centering
    \subfloat[AUC vs. Number of Parameters]{\includegraphics[width=0.8\linewidth]{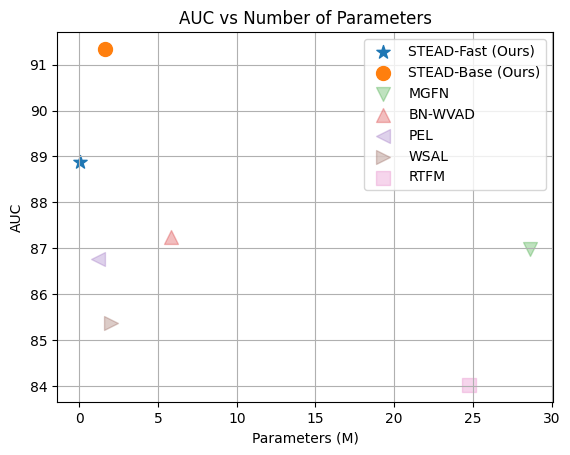}%
    \label{fig:auccompare_a}}
    \vfil
    \subfloat[AUC vs. Number of Parameters (with Feature Extractor)]{\includegraphics[width=0.8\linewidth]{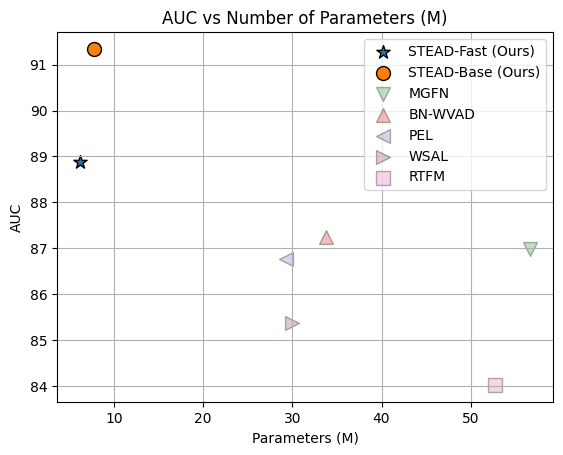}%
    \label{fig:auccompare_b}}
    \caption{Comparison of model performance on the benchmark against model size. (a) Only model size is considered. (b) Feature extractor size is considered alongside model size.}
    \label{fig:auccompare}
\end{figure}

As shown in Table \ref{table:ucf-results}, our models (\emph{full-size version} and \emph{fast version}) achieve state-of-the-art performance, surpassing all previous works by a significant margin. Our base model STEAD-Base achieves an AUC of 91.34\%, outperforming BN-WVAD \cite{zhou2023batchnormbasedweaklysupervisedvideo}, the previous state-of-the-art. Additionally, our fast version STEAD-Fast achieves an AUC of 88.87\%, outperforming the previous state-of-the-art, despite a significant 99.70\% reduction in parameters. The success of our base model using X3D-RGB features over the same model using I3D features cements the superiority of using the X3D model to extract features over the I3D model. Lastly, our models contain significantly less parameters than previous models, making them extremely valuable for \emph{real-time} inference, especially our Fast model which operates under an extremely small number of parameters. Visual comparisons of our models' and previous works' AUC plotted against the total number of parameters in shown in Figure \ref{fig:auccompare}. A visualization of how the state-of-the-art performance has changed over time can be seen in Figure \ref{fig:auc_time}, where it can be observed that improvements have stagnated for the last few years until the completion of our proposed model. Both STEAD-BASE and STEAD-Fast achieve significantly better performance while having similar or less parameter counts than previous works. Factoring in the feature extractor size, our models have a substantial advantage in real-time inference.

\begin{figure}[!t]
    \centering
    \subfloat[Embeddings with Triplet Loss]{\includegraphics[width=0.8\linewidth]{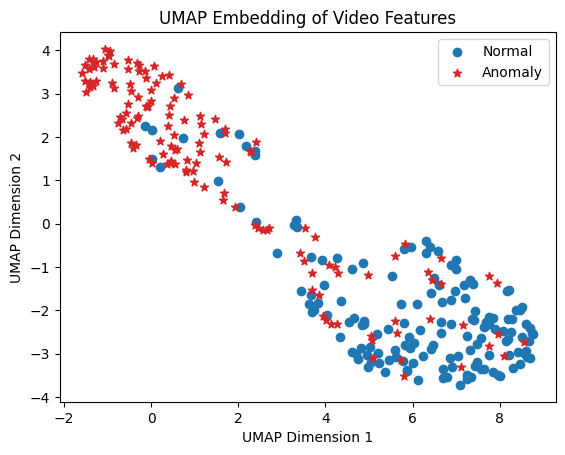}%
    \label{fig:embed_a}}
    \vfil
    \subfloat[Embeddings without Triplet Loss]{\includegraphics[width=0.8\linewidth]{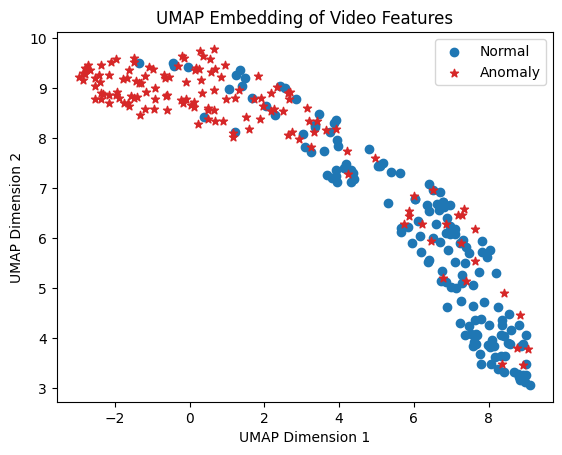}
    \label{fig:embed_b}}
    \caption{A UMAP visualization of the final video feature embeddings using STEAD-Fast. (a) Embeddings created from training the model with triplet loss. (b) Embeddings created from training the model without triplet loss. The distinction between normal and anomaly clusters is much more apparent when using triplet loss.}
    \label{fig:embed}
\end{figure}

\begin{table}[!t]{\small
\begin{center}
\caption{Ablation study on the various elements of STEAD. AUC-ROC results obtained from the evaluating STEAD-Fast on the benchmark. It is shown that the removal of elements results in the degradation of model performance.}
\begin{tabular}{ccc}
\toprule
Ablation & AUC-ROC(\%) & \# of Parameters\\
\midrule
None & 88.87 & 17,441\\
Triplet Loss & 88.12 & 17,441\\
Attention Block & 87.06 & 10,977\\
Feature Enhancer Block & 88.16 & 14,241\\
Both Blocks & 85.90 & 6,657\\
\bottomrule
\end{tabular}

\vspace*{-4mm}
\label{table:ablation}
\end{center}}
\end{table}




We also conduct multiple experiments regarding the ablation of the following elements of our model: \emph{triplet loss}, \emph{Decoupled Spatio-Temporal Feature Enhancer Block}, and \emph{Spatio-Temporal Attention Block}. 

AUC-ROC performance with the ablation of the various components on the benchmark are shown in Table \ref{table:ablation}. A UMAP \cite{mcinnes2020umapuniformmanifoldapproximation} of the final video embeddings is shown in Figure \ref{fig:embed} with embeddings modeled using triplet loss shown in figure \ref{fig:embed_a} and without triplet loss in Figure \ref{fig:embed_b}. STEAD-Fast with triplet loss outperforms its counterpart without triplet loss on the benchmark. Additionally, the distinction between the groups of normal and anomalous embeddings is much clearer with triplet loss than without, showing the effectiveness of triplet loss. 

While ablating the backbone blocks of the model results in reduced performance, the results obtained still compete with previous state-of-the-arts while requiring even less parameters than STEAD-Fast. 
However, while the reduction of parameters when compared against each other is substantial, it is insignificant when considering real-time inference with the addition of 6.15M parameters from the X3D feature extractor.


\section{Conclusion and Future Work}
This paper presents a new method in detecting abnormal behavior in autonomous driving with an emphasis on practicality for real-time inference called STEAD, with efficient decoupled spatiotemporal convolutions and linear approximations of coupled spatiotemporal attention mechanisms. STEAD and its variants achieve state-of-the-art performance, significantly outperforming all previous works with drastically lower model sizes. As an integral element in the transfer learning of the STEAD, the X3D feature extractor is proven to be vastly superior in terms of its effectiveness and efficiency than other feature extractors used by previous works, such as I3D. 

In terms of future work, we hope our paper encourages future works to place greater emphasis on efficiency as we have shown that greater performance does not necessarily need to be a result of sacrificing efficiency.








\bibliographystyle{IEEEtran}
\bibliography{main}

\end{document}